\documentclass[runningheads]{llncs}
\usepackage[T1]{fontenc}
\usepackage{graphicx}
\usepackage{amssymb}  
\usepackage{xcolor}  
\usepackage{orcidlink}
\usepackage[font=small]{caption}
\usepackage{subcaption}

\usepackage{cite}
\usepackage{algorithm}     
\usepackage{graphicx}
\usepackage{booktabs}
\usepackage{xurl}

\usepackage{siunitx}

\usepackage[noend]{algpseudocode} 
\usepackage{amsmath}
\usepackage{cleveref}
\usepackage{multirow}
\usepackage[inline]{enumitem}
\DeclareMathOperator*{\argmin}{arg\,min}

\begin{document}
\title{Relay-Based Coordination for Energy-Efficient Multi-Robot Pickup and Delivery}
\titlerunning{VCST-RCP: Relay-based Coordination for Multi-Robot Delivery}
%
\author{Alkesh K. Srivastava\inst{1}\orcidlink{0000-0002-7470-4620} \and
Jared Michael Levin\inst{1,2}\orcidlink{0009-0005-4022-0053} \and
Philip Dames\inst{1}\orcidlink{0000-0002-7257-0075}}
\authorrunning{A.K. Srivastava et al.}
\institute{
Temple University, Philadelphia, PA, 19122 USA\\
\email{\{alkesh, pdames\}@temple.edu}
\and
Brock Solutions, Kitchener, ON, Canada\\
\email{jlevin@brocksolutions.com}
}
\maketitle              
\begin{abstract}
We consider the problem of delivering multiple packages from a single depot to distinct goal locations using a homogeneous fleet of robots with finite multi-package carrying capacity. We propose VCST-RCP, a Voronoi-Constrained Steiner Tree Relay Coordination Planning framework that explicitly treats inter-robot relays as a design primitive. The approach operates in two stages: (i) constructing a sparse relay backbone by combining Voronoi-derived exchange interfaces with Steiner tree optimization, and (ii) synthesizing robot-level pickup, relay, and delivery schedules under capacity and service-time constraints. Unlike traditional methods that rely on direct source-to-destination transport, our framework {organizes package flow through a shared relay network, reducing redundant long-haul motion}. Extensive experiments across multiple scales show that VCST-RCP reduces total fleet travel distance by an average of 31\% (up to nearly 50\%) compared to Hungarian assignment and significantly outperforms OR-Tools CVRP, with statistically significant improvements ($p < 10^{-3}$). These gains translate into over 50\% higher delivery efficiency (packages per kilometer), directly improving energy utilization. An ablation study further reveals that optimizing relay placement yields substantially larger improvements than adapting spatial partitioning alone, establishing relay design as the dominant factor governing system performance. Overall, the results demonstrate that relay-based coordination provides a scalable and effective framework for energy-aware multi-robot delivery in real-world logistics settings.

\end{abstract}

\section{Introduction}

\begin{figure}[t]
    \centering
    \includegraphics[width=0.7\textwidth]{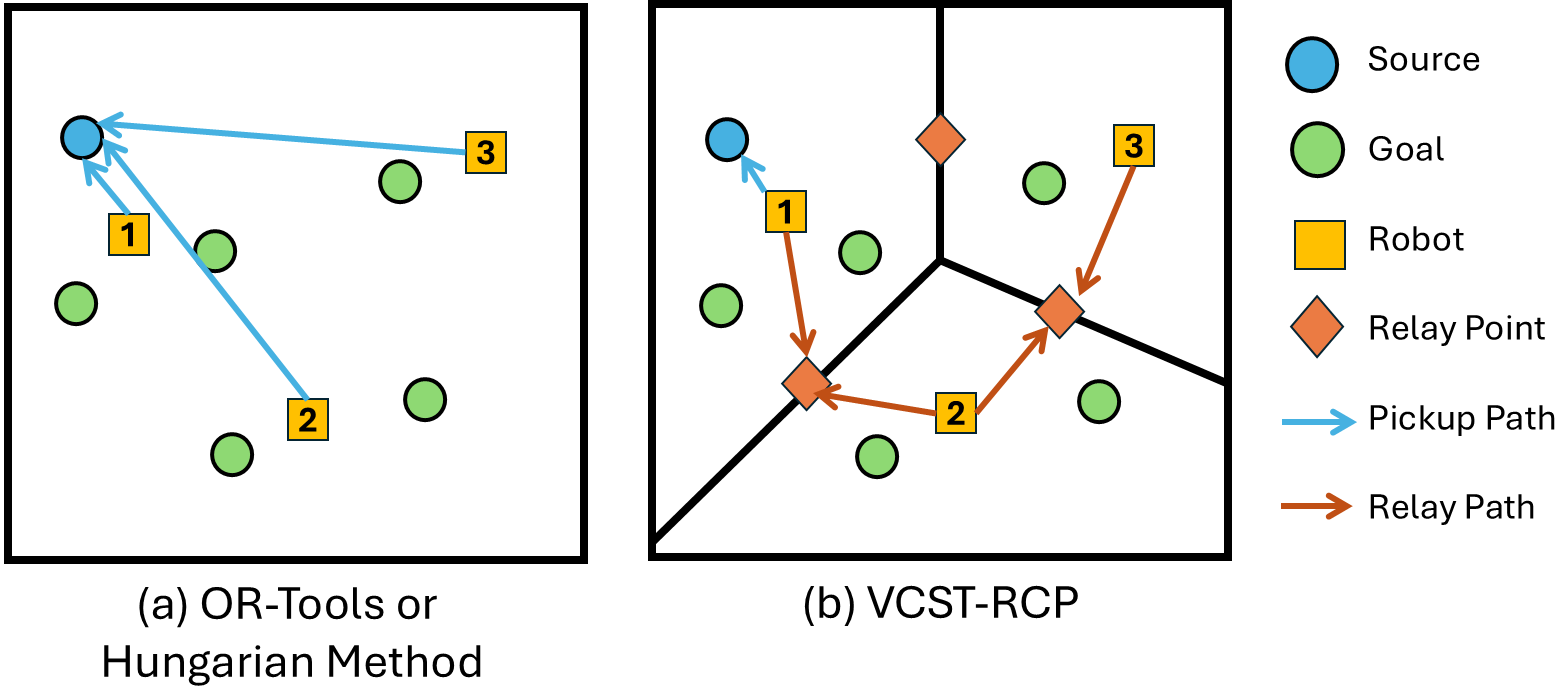}
    \caption{Conceptual comparison of classical planners and our approach. 
    (a) Capacitated Vehicle Routing Problem (CVRP) or Hungarian assignment dispatch robots on direct source-to-goal trips, causing redundant long-haul travel. 
    (b) VCST-RCP constructs Voronoi-constrained Steiner trunks with relay points, concentrating motion and reducing redundancy. 
    Robots then continue from relays to goals (not shown for clarity).}
    \label{fig:concept}
\end{figure}

Mobile robot fleets are increasingly used to transport goods across campuses, warehouses, and urban environments. Starship Technologies, for example, operates thousands of sidewalk delivery robots that collectively travel tens of kilometers per robot per day~\cite{Jordan_2023}. Since locomotion dominates the energy consumption of ground robots~\cite{wu2023review}, fleet travel distance serves as a practical proxy for energy expenditure. Classical planners such as the Hungarian assignment algorithm~\cite{kuhn1955hungarian} and the Capacitated Vehicle Routing Problem (CVRP)~\cite{ralphs2003capacitated} optimize direct {delivery without allowing packages to transfer between robots during execution. However, fleets may reuse portions of shared transport structure across multiple package flows and thereby reduce total fleet travel} and improve delivery efficiency. \Cref{fig:concept} illustrates how direct source-to-goal transport leads to redundant long-haul motion, whereas relay-based coordination can concentrate package flow.

Relays are already central to large-scale logistics. Amazon, for instance, has deployed more than 750,000 robots in fulfillment operations where packages move through fixed transfer stations and robotic workcells before downstream processing~\cite{Greenawalt_2024}. Such structured depots simplify coordination, but they also limit flexibility because relay sites are predefined rather than optimized. In contrast, campus- and city-scale deployments operate in open environments where exchanges could occur opportunistically at intersections, sidewalks, or hallways. What remains missing is a principled framework for selecting and coordinating relay locations in open environments. Our work addresses this gap by treating relays not as incidental handoffs, but as a design primitive for multi-robot delivery.

We propose \emph{Voronoi-Constrained Steiner Tree Relay Coordination Planning} (VCST-RCP), a relay-based framework for energy-aware multi-robot delivery. The key idea is to move beyond direct source-to-goal routing and instead coordinate robots through a shared relay backbone that explicitly structures how packages flow through the team. By enabling coordinated short-range exchanges, VCST-RCP reduces redundant long-haul motion and improves overall transport efficiency. Our results further show that these gains are primarily driven by relay design, establishing relay structure as the dominant factor governing transport efficiency.

\textbf{Contributions}: This paper makes the following contributions.

\textbf{(i) Relay-centric transport network design:}
We introduce VCST-RCP, a coordination framework that elevates relays from an incidental mechanism to a principled design primitive for multi-robot delivery. 
The proposed \emph{Steiner Relay Trunk}, constructed via geometric relay selection and network optimization, reduces redundant long-haul motion across the fleet. Conceptually, this induces a handoff-based transport mechanism analogous to a ``bucket brigade,'' where packages propagate through coordinated short-range exchanges rather than independent long-haul trips.

\textbf{(ii) Capacity- and service-aware relay execution:}
We develop a two-stage planning framework that converts the relay backbone into executable robot schedules. Stage~1 constructs the Steiner relay trunk using Voronoi-constrained candidate relay locations and graph-theoretic optimization~\cite{hwang1992steiner}. Stage~2 compiles robot-level pickup, relay, and delivery timelines that explicitly account for robot capacity limits and non-zero service times, translating network flows into coordinated robot actions.

\textbf{(iii) Empirical evaluation and analysis of relay-based coordination:}
We present a large-scale empirical study of relay-based multi-robot delivery under capacity constraints, demonstrating statistically significant reductions in total fleet travel distance of \textbf{31\% on average (up to nearly 50\%)} compared to Hungarian assignment, and consistent improvements over OR-Tools CVRP, along with over \textbf{50\% higher delivery efficiency} (packages per kilometer). Our analysis further shows that optimizing relay placement yields substantially larger gains than adapting spatial partitioning alone, underscoring the central role of relay design in multi-robot transport efficiency.

\section{Related Work}

Classical routing problems such as the Traveling Salesman Problem (TSP), its multi-agent variant (mTSP), and Vehicle Routing Problems (VRP) provide the foundations for distance-minimizing logistics and fleet coordination~\cite{junger1995traveling,bektas2006multiple,toth2002vehicle,eksioglu2009vehicle,ortools_cvrp}. These approaches model assignment and capacity constraints effectively, but typically assume direct source-to-destination transport and do not explicitly consider relay-based exchanges between robots.

Multi-Agent Path Finding (MAPF) and Multi-Agent Pickup and Delivery (MAPD) address collision-free coordination, task assignment, and execution for robot teams with evolving task sets~\cite{sharon2015conflict,sharon2013increasing,surynek2016efficient,yu2013multi,gao2024review,salzman2020research,liu2019task,ma2017lifelong}. Rolling-horizon methods further improve responsiveness in dynamic environments~\cite{song2016rolling,wang2015rolling}. However, these formulations primarily optimize who moves where and when under direct transport assumptions, rather than designing a shared relay backbone that structures package flow through the environment.

\begin{figure}[t]
  \centering
  \includegraphics[width=0.8\textwidth]{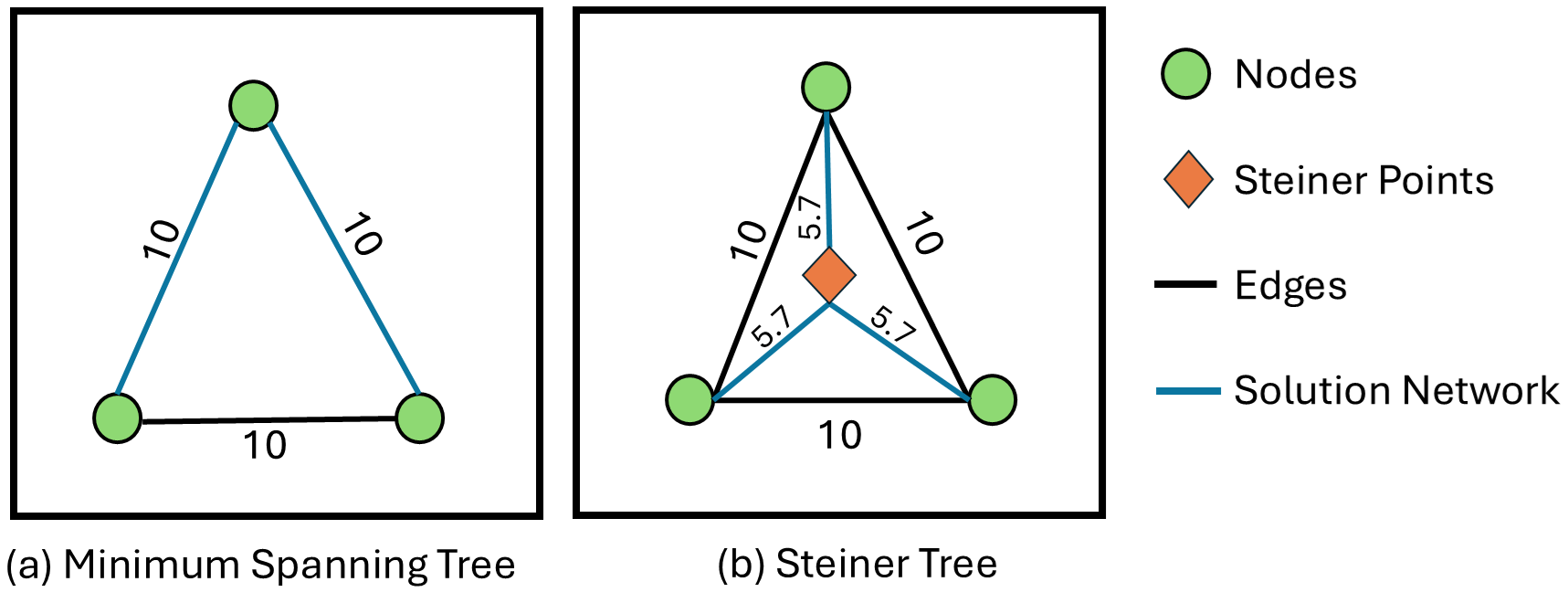}
  \caption{Minimum Spanning Tree vs.\ Steiner tree for three terminals forming an equilateral triangle (side length \(10\)). In each panel the \textbf{solution network} is shown in blue, while remaining candidate edges are gray. (a) MST uses two edges (total \(=20\)). (b) Adding a Steiner point yields three spokes of length \(10/\sqrt{3}\approx 5.77\) each (total \(\approx 17.32\)), a \(13.4\%\) reduction {in total connecting-network length}.}
  \label{fig:steiner-vs-mst}
\end{figure}

Steiner trees provide a natural framework for connecting required terminals through optional intermediate nodes at low cost~\cite{hwang1992steiner}. Figure~\ref{fig:steiner-vs-mst} illustrates how introducing a Steiner point can shorten the connecting network relative to a minimum spanning tree. This perspective is well matched to relay selection, where the source and goals act as terminals and candidate relay locations act as Steiner nodes. Separately, Voronoi partitioning has long been used in multi-robot systems for adaptive sampling, coverage, exploration, target tracking, and workspace decomposition~\cite{kemna2017multi,srivastava2022distributed,munir2024anchor,hu2020voronoi,luo2019voronoi,kim2020voronoi,huang2021path,lee2024adaptive,nair2020gm,guruprasad2012distributed,dames2020distributed,bakolas2010zermelo, srivastava2026bayesian}. Most closely related, Voronoi relays have recently been explored for single-package delivery with an emphasis on language-guided coordination and robustness~\cite{srivastava2025deliver}. However, prior work has not combined Voronoi-derived relay interfaces with Steiner-style network design to construct sparse relay trunks for cooperative multi-robot package delivery under capacity and service-time constraints.

Energy has also emerged as a central concern in persistent logistics. Recent MAPD work has considered battery-aware task allocation and charging trade-offs~\cite{fumiya2024anytime}. Our focus is complementary: because propulsion dominates the energy budget of ground robots, we target total fleet travel distance as an energy proxy and study how relay-based transport backbones reduce redundant motion under capacity constraints and non-zero service times.

\section{Problem Formulation}

We consider the problem of delivering $N$ identical packages from a single source location $S$ to a set of goal locations $G=\{g_1,g_2,\ldots,g_N\}$ using a homogeneous team of $M$ robots operating in a bounded, convex workspace $\mathcal{E}\subset\mathbb{R}^2$.  This setting arises naturally in applications such as in a distribution center or last-mile delivery from a restaurant, where robots must transport items from a central depot to geographically dispersed customers~\cite{keith2024review}. 

Each robot $r_m\in R=\{r_1,\ldots,r_M\}$ is defined by an initial position $p_m\in\mathcal{E}$, a maximum carrying capacity $C$, and a fixed speed $v_{\textrm{speed}}$. Robots are capable of package pickup, delivery, and relay at designated relay locations, each of which requires a fixed service time $T_s$. A feasible delivery plan must respect capacity limits, account for these non-zero service times, and ensure that every package departing the source is delivered to a unique goal. In contrast to traditional formulations where routes are fixed between source and goals, our setting also allows the selection of intermediate relay locations as part of the planning problem, enabling coordinated load exchange between robots.


The objective is to compute coordinated multi-robot plans $\Pi = \{\pi_m\}_{m=1}^M$, where $\pi_m$ denotes the plan executed by robot $r_m$, that minimize the total travel distance of the fleet:
\begin{equation}
\label{eq:objective}
    \min_{\Pi} \sum_{m=1}^{M} d(\pi_m),
\end{equation}
{where $d(\pi_m)$ is the distance traveled by robot $r_m$ along plan $\pi_m$. Importantly, this objective measures total robot travel rather than the geometric distance traversed by individual packages. Relay transport may therefore introduce intermediate package transfers while still reducing the total motion accumulated by the fleet. Since locomotion dominates energy usage in ground robots, total travel distance serves as a proxy for fleet-level energy consumption.}

\section{Voronoi-Constrained Steiner Tree Relay Coordination Planning ({VCST-RCP})}

\label{sec:PF}
\label{sec:{VCST-RCP}}
\begin{figure*}[t]
  \centering
  \includegraphics[width=0.9\textwidth]{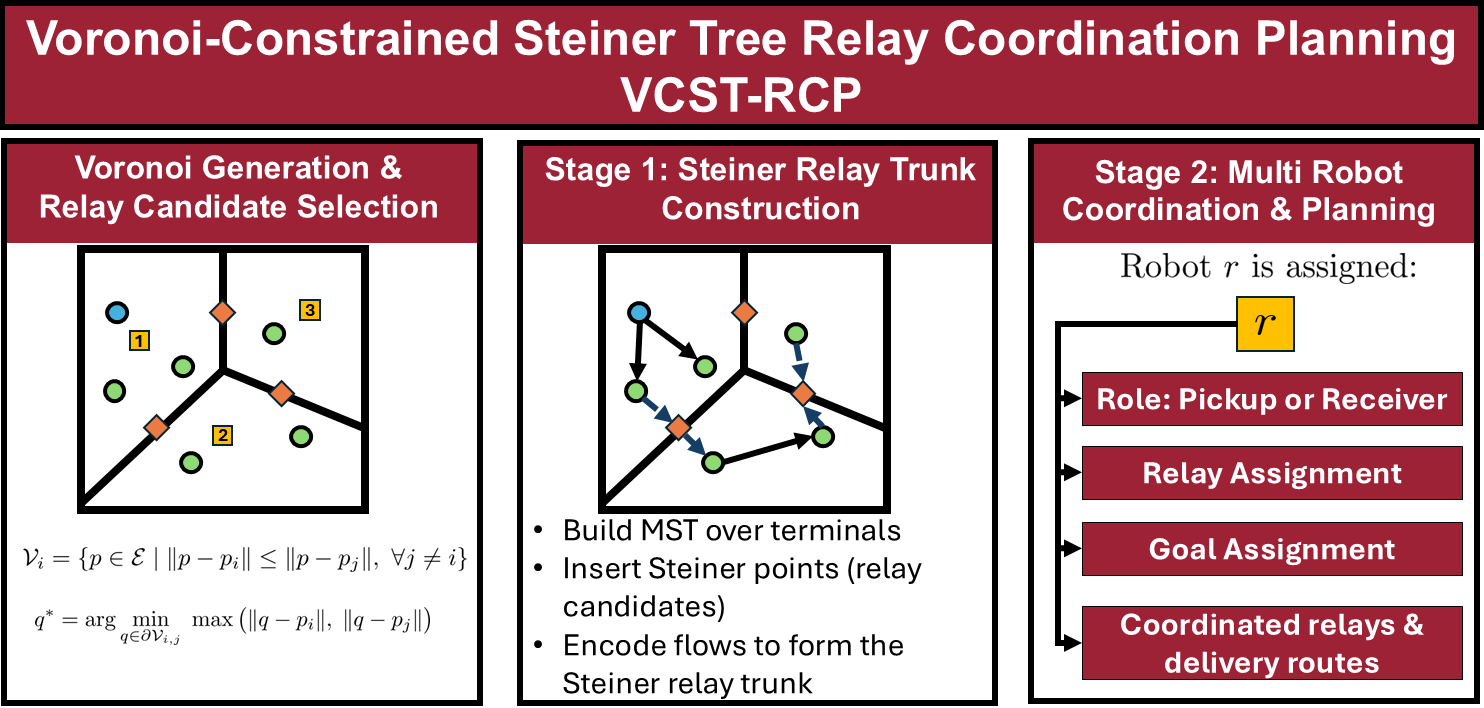}
  \caption{\textbf{VCST-RCP system overview.}
\emph{Left—} Robots induce a Voronoi partitioning of the workspace; each shared boundary yields a relay candidate (orange diamond).
\emph{Middle—Stage 1:} A Steiner relay trunk is built to connect the depot to all goals with minimal travel-time cost.
\emph{Right—Stage 2:} Robots are assigned pickup/receiver roles and goals, and an asynchronous schedule (drop now, pick up later) is compiled under capacity and per-stop service time to minimize completion time.}

  \label{fig:system-overview}
\end{figure*}

We introduce \emph{Voronoi-Constrained Steiner Tree Relay Coordination Planning} (VCST-RCP), a relay-based framework for multi-robot package delivery that explicitly optimizes where robots should exchange loads. As illustrated in Fig.~\ref{fig:system-overview}, VCST-RCP constructs a sparse transport backbone composed of strategically selected relay points that concentrate package flow and reduce redundant long-haul travel. The framework operates in two stages: in Stage~1, candidate relay locations are generated from the geometric structure of the robot team and a Steiner-style network design selects a sparse subset that forms a shared relay backbone from the source to all goals; in Stage~2, this backbone is translated into executable robot behaviors by assigning pickup, relay, and delivery roles, and scheduling their interactions under capacity and service-time constraints.


\subsection{Relay Modeling and Graph Abstraction}
\label{sec:RMGA}

Unlike structured logistics systems such as hub-and-spoke networks (e.g., postal or parcel delivery), where relays occur at fixed depots or sorting centers, we consider open environments (e.g., campus or urban delivery) where exchanges may occur opportunistically. In such settings, restricting relays to predefined locations can lead to inefficient coordination. We therefore consider an unconstrained two-dimensional workspace in which relay locations are not fixed a priori. Instead, candidate relay points are generated from both robot positions and environment geometry, and are treated as decision variables optimized during planning. Notably, our formulation naturally subsumes fixed-depot settings as a special case by including predefined relay nodes in the candidate set.

\paragraph{Voronoi Partitioning:} 
We partition the workspace into Voronoi regions induced by a set of robot positions, thereby defining geometric zones of responsibility. Specifically, the Voronoi cell $\mathcal{V}_i$ associated with robot $r_i$ located at $p_i$ is given by
\begin{equation}
    \mathcal{V}_i = \left\{ p \in \mathcal{E} \;\middle|\; 
    \|p - p_i\| \leq \|p - p_j\|, \ \forall j \neq i \right\}.
\end{equation}
This partition defines pairwise interfaces between neighboring robots that serve as feasible regions for relay interactions.

Different choices of robot positions lead to different partitions. In VCST-Geo, the geometric instantiation of VCST-RCP, the initial robot positions are used directly to construct the Voronoi diagram. Alternatively, the partition can be adapted to the task by refining robot positions prior to partitioning. In particular, we consider a Lloyd~\cite{bullo2009distributed} algorithm-based variant in which robot positions are iteratively updated toward demand-weighted centroids derived from a Gaussian Mixture Model (GMM) fit to the goal locations. This produces a demand-aware partition while leaving the remainder of the VCST-RCP pipeline unchanged. We compare these variants in the experimental evaluation to isolate the impact of partition design.
\paragraph{Relay Point Computation:}
A Voronoi boundary $\partial\mathcal{V}_{i,j}$, shared by robots $r_i$ and $r_j$, defines a feasible interface for coordinated load exchange, ensuring that candidate relay locations lie in regions accessible to both robots. More generally, relay selection can be formulated as choosing a point $q \in \partial\mathcal{V}_{i,j}$ that minimizes a task-dependent cost function,
\begin{equation}
\label{eq:relaypoint}
    q^*_{ij} = \argmin_{q \in \partial\mathcal{V}_{i,j}} \; \mathcal{C}(q; p_i, p_j),
\end{equation}
where $\mathcal{C}(\cdot)$ encodes the cost of establishing a relay at location $q$.

In this work, we adopt a geometric formulation as the canonical choice, where the cost function balances the travel effort of the two participating robots:
\begin{equation}
\label{eq:relaycost}
    \mathcal{C}(q; p_i, p_j) = \max \big( \|q - p_i\|,\; \|q - p_j\| \big).
\end{equation}
This yields relay locations that minimize the maximum travel required for either robot to participate in the exchange, promoting efficient and balanced coordination. This formulation relies only on robot geometry and does not incorporate information about the distribution of sources or goals, and therefore serves as a natural baseline for evaluating the impact of relay design. In practice, the optimal solution is given by the intersection of the perpendicular bisector of $p_i$ and $p_j$ with $\partial\mathcal{V}_{i,j}$, or by the nearest endpoint of the Voronoi edge if the bisector does not intersect within the segment.

This formulation also naturally accommodates extensions where relay selection incorporates task-specific information. In the demand-aware variant, a Gaussian Mixture Model (GMM) is fitted to the goal locations, yielding component means $\{\mu_k\}_{k=1}^K$ and weights $\{w_k\}_{k=1}^K$. For each Voronoi boundary edge $\partial V_{ij}$ shared by robots $i$ and $j$, the relay candidate is selected as
\begin{equation}
    q^*_{ij} = \argmin_{q \in \partial V_{ij}}
    \sum_{k=1}^{K} w_k \, \|q - \mu_k\|,
    \label{eq:gmm_relay}
\end{equation}
where the minimization is performed by uniform line search along $\partial V_{ij}$. This biases relay locations toward regions of concentrated delivery demand while preserving the geometric constraint that relays lie on shared Voronoi interfaces. The partition itself remains unchanged; only relay placement along each boundary is modified. We evaluate the impact of this demand-aware relay selection in~\Cref{sec:experiments}.


\paragraph{Graph-Theoretic Abstraction:}
We represent the relay-based transport structure as a graph $\mathcal{G} = (V,E)$, where the vertex set $V = \{S\} \cup G \cup Q$ consists of the (single) source location $S$, all goal locations $G$, and a set of candidate relay locations $Q = \{q_{ij}^* \mid i \neq j\}$ (from \eqref{eq:relaypoint}). Edges $(u,v) \in E$ represent feasible travel between locations, weighted by a cost that captures both travel distance and service-time penalties.

The candidate relay set $Q$ is constructed from pairwise interfaces between robots. For each pair of adjacent Voronoi cells $(\mathcal{V}_i, \mathcal{V}_j)$, we compute a relay location $q^*_{i,j}$ along the shared boundary using~\eqref{eq:relaypoint}, and include it in $Q$. These points define feasible locations where coordinated load exchange can occur, and serve as candidate nodes for shaping the transport structure.




{VCST-RCP} operates in two stages: (i) building a Steiner tree that connects the source $S$ and all goals $G$ while optionally introducing relay nodes $Q$, which we interpret as a \emph{Steiner Relay Trunk}~(\Cref{def:SRT}), the strategic backbone of package flow, and (ii) planning robot-level execution of pickups, relays, and deliveries along this trunk. In our setting, this backbone defines where coordination occurs and how transport effort is distributed across the robot team. While the resulting structure is sparse and efficient, it reflects a design choice that prioritizes coordinated relay over redundant or highly branched routing structures.

\begin{definition}[Steiner Relay Trunk]
\label{def:SRT}
Given a Steiner tree $\mathcal{T}\subseteq\mathcal{G}$ spanning the source $S$ and all goals $G$, with candidate relays $Q$ as Steiner points, we define the \emph{relay trunk} as the sparse subgraph $\mathcal{T}$ interpreted as a package propagation backbone. 
Unlike a purely graph-theoretic Steiner tree, the relay trunk is endowed with flow semantics. Each edge is interpreted as a directed package transfer route from the source toward a goal, with an associated travel cost proportional to distance and robot velocity. Nodes are typed as source, relay, or goal, where relay nodes serve as transfer sites that incur service times for handoffs and can model capacity-limited queues. This transforms the Steiner tree from a static cost-minimizing structure into a flow-aware backbone.
\end{definition}

\subsection{Stage 1 - Strategic Network Planning}
The first stage of {VCST-RCP} constructs the Steiner relay trunk $\mathcal{T}$, which serves as the strategic backbone of package flow. We compute $\mathcal{T}$ on the graph $\mathcal{G}=(V,E)$ using edge weights
\begin{equation}
\label{eq:cost}
c(u,w)\;=\;\frac{\|u-w\|}{v_{\mathrm{speed}}}\;+\;\lambda_{\mathrm{svc}}\,\sigma(u,w),
\end{equation}
where $\|u-w\|$ is the Euclidean distance between nodes $u, w \in V$, $v_{\mathrm{speed}}$ is the uniform robot speed, $\sigma(u,w)$ denotes the total service time incurred along edge $(u,w)$ (e.g., due to pick-ups, relays, or deliveries), and $\lambda_{\mathrm{svc}}$ is a weighting parameter that balances travel time against service costs.

{We adopt the standard \emph{metric-closure/MST} Steiner heuristic~\cite{hwang1992steiner}, which provides a polynomial-time approximation to the minimum-cost Steiner network under the graph edge metric.} The method computes all-pairs shortest-path costs among terminals $T=\{S\}\cup G$ to form the metric closure, builds a minimum spanning tree (MST) on this closure, and then expands each MST edge back into its corresponding shortest path in the original graph $\mathcal{G}$. The union of these expanded paths yields a sparse relay trunk that approximates the minimum-cost Steiner tree while remaining computationally tractable for large problem instances.

\begin{figure}[t]
  \centering
  \includegraphics[width=0.6\textwidth]{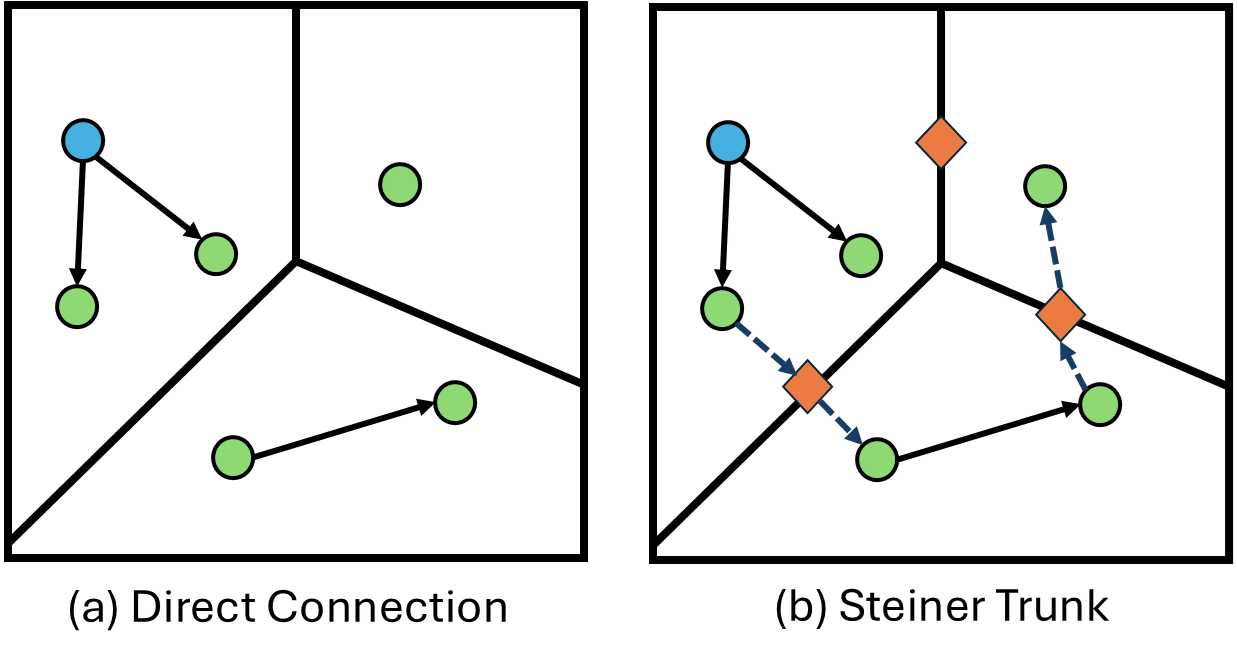}
  \caption{Illustration of Steiner Trunk Construction (Stage~1). 
  (a) Terminals (source in blue and goals in green) are initialized and direct shortest paths are considered. 
  (b) A Steiner trunk is formed by computing an MST over terminals, expanding edges into shortest paths in the environment graph, and introducing relay nodes (orange). Flows are then routed along this trunk, concentrating motion on a sparse backbone that serves all goals.}
  \label{fig:steiner-trunk}
\end{figure}

Each goal $g \in G$ induces a demand from the source $S$, which may be uniform or vary across goals depending on the task. Routing these demands along $\mathcal{T}$ induces integer flows $f(u,v)$ on edges $(u,v) \in \mathcal{T}$, which quantify the number of packages carried across each relay segment. These flows directly inform the coordination stage by indicating where robot interactions (relays) must occur. 
The complete procedure for constructing $\mathcal{T}$ and computing $f(u,v)$ is illustrated in~\Cref{fig:steiner-trunk} and summarized in~\Cref{alg:steiner-trunk}.

\begin{algorithm}[t]
\caption{Steiner Trunk Construction (Stage~1)}
\label{alg:steiner-trunk}
\begin{algorithmic}[1]
\State \textbf{Input:} Source $S$, goals $G$, candidate relay points $H$, graph $\mathcal{G}=(V,E)$
\State \textbf{Output:} Steiner relay trunk $\mathcal{T}$, edge flows $f(u,v)$
\State Initialize terminals $T \gets \{S\} \cup G$
\State Construct complete graph $K_T$ on $T$ with edge weights equal to the shortest-path costs in $\mathcal{G}$ (measured by cost metric $c$, e.g., Euclidean distance)
\State Compute minimum spanning tree $\mathrm{MST}(K_T)$
\State For each edge $(t,t')\in \mathrm{MST}(K_T)$, retrieve a shortest path $P_{t\to t'}$ in $\mathcal{G}$ (under $c$)
\State Form trunk $\mathcal{T} \gets \bigcup_{(t,t')\in \mathrm{MST}(K_T)} P_{t\to t'}$ (introducing relay nodes from $H$ when traversed)
\ForAll{$g \in G$}
    \State Route one unit of demand from $S$ to $g$ along its path in $\mathcal{T}$ and accumulate edge flows $f(u,v)$
\EndFor
\State \textbf{Return:} Trunk $\mathcal{T}$ and flows $f(u,v)$
\end{algorithmic}
\end{algorithm}


\subsection{Stage 2 - Robot Coordination Planning}

\begin{algorithm}[t]
\caption{Robot Coordination Planning (Stage~2)}
\label{alg:stage2}
\begin{algorithmic}[1]
\State \textbf{Input:} Relay trunk $\mathcal{T}$ with flows $f(u,v)$, robot poses, speed $v_{\mathrm{speed}}$, capacity $C$, service time $T_s$
\State \textbf{Output:} Executable per-robot timelines

\State Aggregate relay demand at each node using flows $f(u,v)$
\State Select pickup robot at source $S$

\While{unsatisfied relay demand exists}
    \State Batch up to $C$ packages
    \State Construct pickup tour from $S$ to required relay nodes
    \State Deposit packages at relays and record availability times
\EndWhile

\State Assign relay nodes to receiver robots based on proximity and workload

\ForAll{goals $g \in G$}
    \State Determine final relay node along trunk path
    \State Assign goal delivery to corresponding robot
\EndFor

\ForAll{robots}
    \State Construct delivery route using MST-preorder heuristic
\EndFor

\State Generate robot timelines with \texttt{travel}, \texttt{deposit}, \texttt{pickup}, and \texttt{deliver} actions
\State Enforce service times and package availability constraints
\State \textbf{Return:} executable timelines
\end{algorithmic}
\end{algorithm}

In the planning phase, we take the Steiner relay trunk $\mathcal{T}$ and its edge flows $f(u,w)$ from Stage~1 and produce an executable multi-robot plan that respects capacity and service times while realizing the prescribed package propagation. The inputs are: (i) trunk $\mathcal{T}$ and per-edge flows $f(u,w)$, (ii) robot initial poses, shared speed $v_{\mathrm{speed}}$, and package capacity $C$, and (iii) uniform service time $T_s$ for pickup, relay, and delivery. The output is a set of per-robot timelines containing \texttt{travel}, \texttt{relay}, and \texttt{deliver} actions with timestamps

\paragraph{Pickup Batching:} We first determine the package demand at each relay node by aggregating flows from $\mathcal{T}$. The designated pickup robot at $S$ executes a sequence of batched tours, each constrained by capacity $C$, to deliver packages to relay nodes. Each visit incurs a service time $T_s$, and multiple tours may be required to satisfy demand.

\paragraph{Relay Assignment:}
Given the flow values $f(u,v)$ on the trunk $\mathcal{T}$, each relay edge $(u,v)$ induces a transport requirement corresponding to the number of packages that must traverse that segment. Since relay locations lie on pairwise Voronoi interfaces, each relay node is naturally associated with a pair of neighboring robots. We assign relay responsibilities by mapping each relay node to one of its adjacent robots, prioritizing proximity and balancing workload across the team. This step translates edge-level flows into robot-level relay responsibilities.

\paragraph{Delivery Planning:}
For each goal $g \in G$, we identify its path from the source $S$ along the trunk $\mathcal{T}$. If the path contains relay nodes, the package is handed off sequentially along this path; otherwise, it is delivered directly by the pickup robot. Goals assigned to the same robot are grouped based on their corresponding terminal segments of $\mathcal{T}$, and delivery routes are constructed by traversing these segments. This results in structured, tree-constrained delivery paths without requiring the solution of a general TSP. 

\paragraph{Timeline Synthesis:}
Finally, we construct executable per-robot timelines by sequencing pickup, relay, and delivery actions along the paths induced by $\mathcal{T}$. For each robot, tasks are ordered according to their position along the trunk and the direction of flow, ensuring that upstream pickups precede downstream relays and deliveries. Relays are modeled asynchronously\footnote{Future work will incorporate an explicit synchronization constraint requiring co-presence at relays.}: a robot arriving at a relay location deposits packages, and the receiving robot collects them once available. If a robot arrives before the corresponding package is available, it waits until the preceding transfer is completed. Each action is annotated with service duration $T_s$, and travel times are determined by edge costs.

This construction yields feasible execution schedules that respect capacity limits and temporal dependencies induced by relay ordering, thereby transforming the abstract flows on $\mathcal{T}$ into coordinated robot behaviors.

\section{Experiments and Results}
\label{sec:experiments}

We evaluate VCST-RCP across a suite of delivery environments designed to capture variations in spatial scale, demand distribution, and capacity constraints. The scenarios include small dense and sparse settings, medium-scale balanced and high-capacity environments, large warehouse and distribution layouts, and capacity-constrained regimes. Each experiment is repeated over 100 randomized trials, and we report mean performance with standard deviation. Statistical significance is indicated using standard notation ($^{*}p<0.05$, $^{**}p<0.01$, $^{***}p<0.001$).

\subsection{Baseline Methods}

We compare against two standard baselines: (i) Hungarian assignment, which performs direct source-to-goal matching without coordination, and (ii) OR-Tools CVRP, which computes capacitated vehicle routes using optimized tours. In addition, we include a geometric instantiation of our framework (VCST-Geo), which uses purely geometry-driven relay selection without incorporating task or demand information. This variant serves as a reference to isolate the impact of relay-aware design within VCST-RCP.

\subsection{Results}

\begin{figure}[t]
    \centering
    \includegraphics[width=\textwidth]{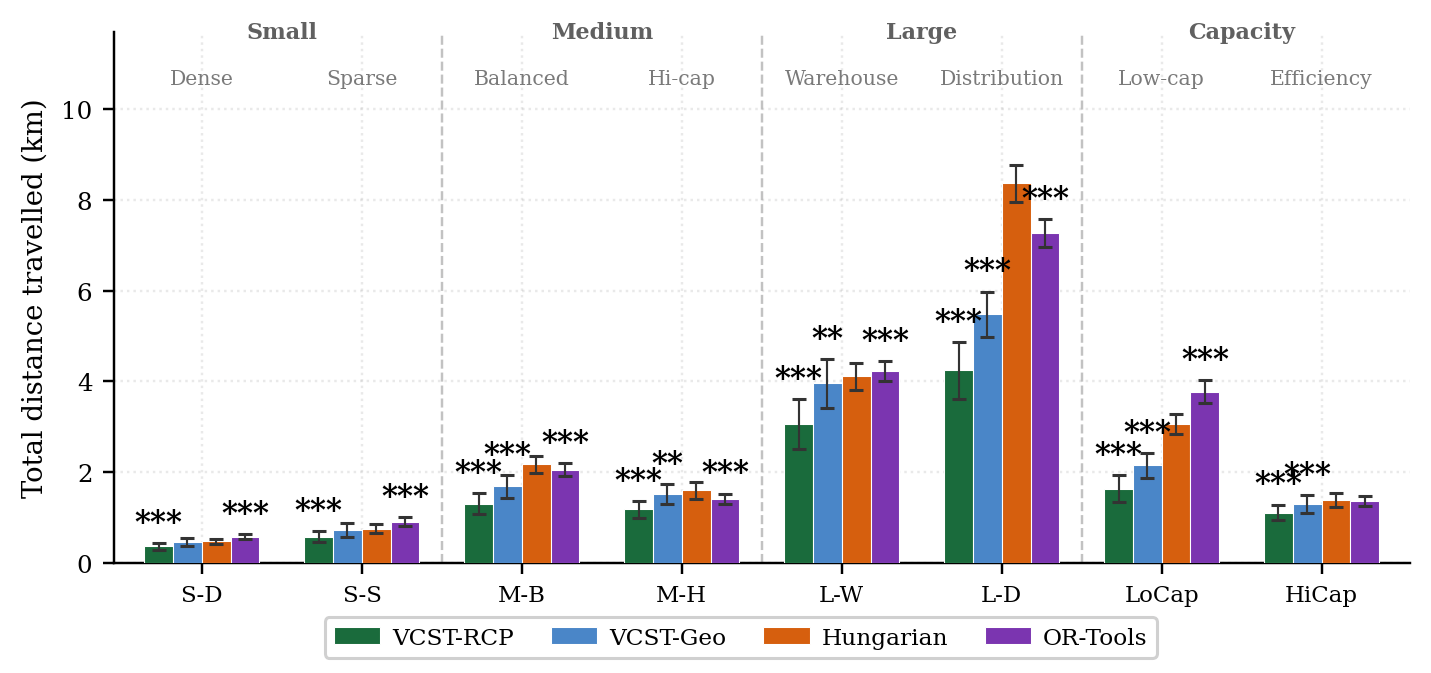}
    \caption{Total fleet travel distance (km) across scenarios (mean $\pm$ std). Environments are grouped by size (Small, Medium, Large) and capacity regimes. VCST-RCP consistently achieves the lowest travel distance across all settings, with statistically significant improvements over all baselines.}
    \label{fig:distance}
\end{figure}

\noindent\textbf{(i) Travel Distance:} Figure~\ref{fig:distance} shows total fleet travel distance across all scenarios. VCST-RCP consistently outperforms all baselines, achieving the lowest travel cost in every environment. Compared to Hungarian assignment, VCST-RCP reduces total distance by an average of approximately \textbf{31\%}, with improvements ranging from \textbf{+18\% to nearly +50\%} depending on the scenario. The gains are especially pronounced in large-scale environments, where long-haul transport dominates. In these settings, relay-based coordination significantly reduces redundant motion by concentrating package flow through shared transport paths. While the geometric variant (VCST-Geo) provides moderate improvements, it is consistently outperformed by VCST-RCP, highlighting the importance of relay-aware optimization.

\begin{figure}[t]
    \centering
    \includegraphics[width=\textwidth]{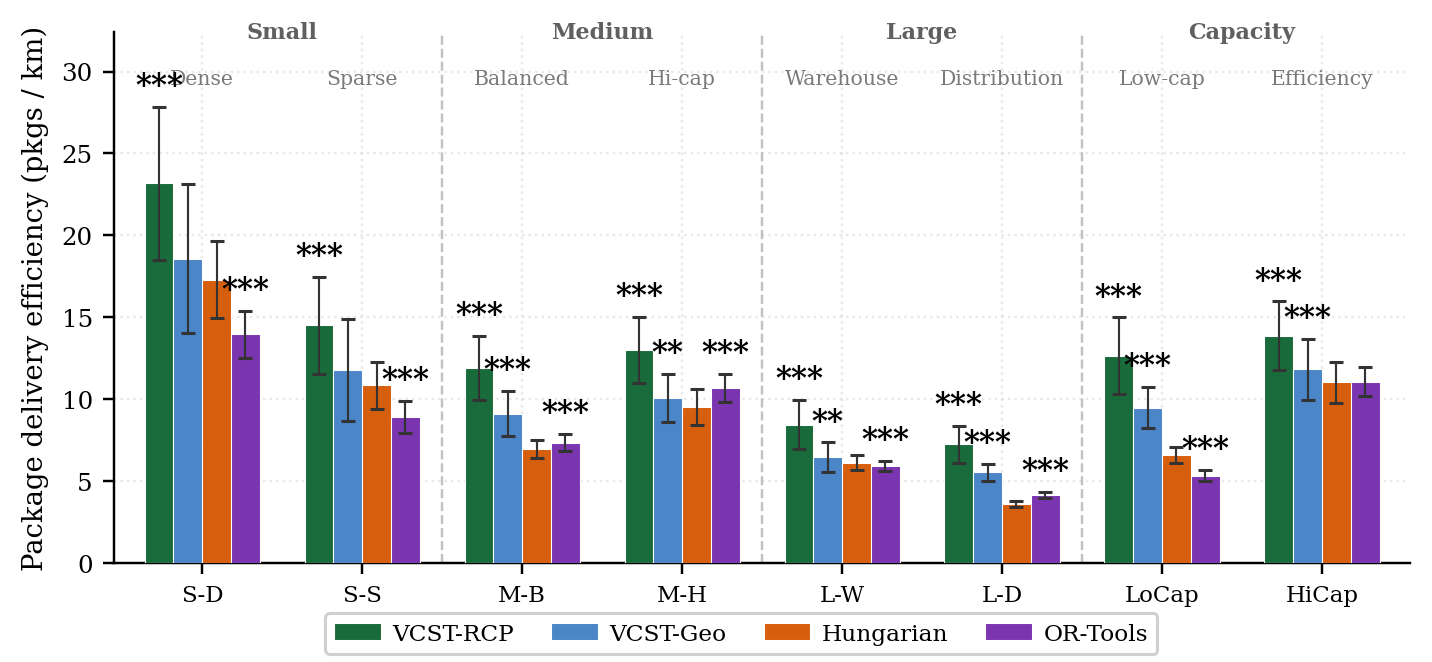}
    \caption{Delivery efficiency (packages per km) across scenarios (mean $\pm$ std). VCST-RCP achieves significantly higher throughput per unit distance, demonstrating improved utilization of robot motion.}
    \label{fig:efficiency}
\end{figure}

\noindent\textbf{(ii) Delivery Efficiency:}
To further evaluate system performance, we measure delivery efficiency in terms of packages delivered per unit distance. As shown in Fig.~\ref{fig:efficiency}, VCST-RCP achieves substantial improvements in efficiency across all scenarios, with an average increase of over \textbf{50\%} relative to Hungarian assignment. These results indicate that VCST-RCP not only reduces total travel cost but also improves how effectively robot motion is utilized. In contrast, OR-Tools often underperforms in these settings due to inefficient routing under distributed demand, and the geometric variant again provides intermediate performance.

\noindent\textbf{(iii) Ablation Study (Relay vs. Partition Design):}
We conduct an ablation study to isolate the contributions of relay design and partition adaptation. Specifically, we compare: \begin{enumerate*}
    \item[(i)] a geometric instantiation of VCST,
    \item[(ii)] a partition-adaptive variant (GMM-Lloyd), and 
    \item[(iii)] the full relay-adaptive VCST-RCP.
\end{enumerate*}

\begin{figure}[!t]
    \centering
    \includegraphics[width=\textwidth]{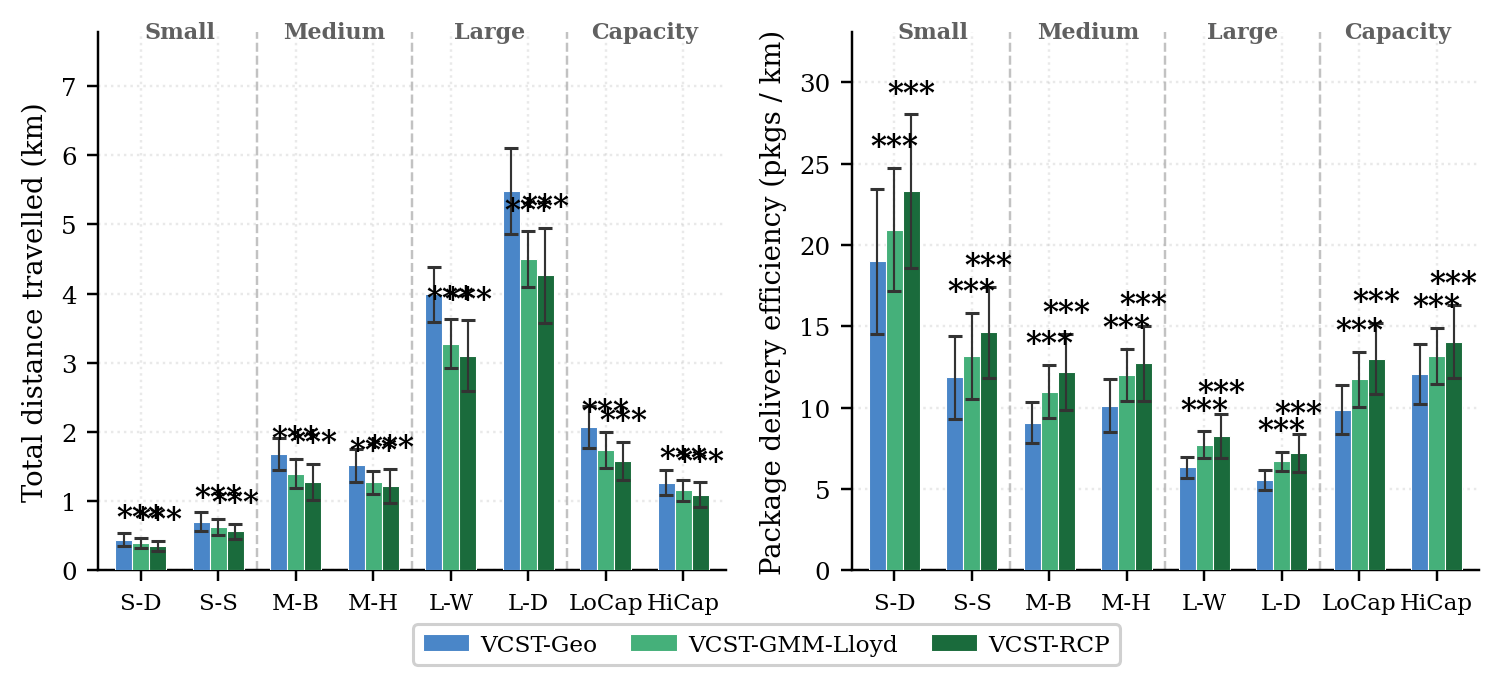}
    \caption{Ablation study comparing relay-aware design (VCST-RCP) and partition-aware design (GMM-Lloyd) against the geometric baseline. Relay optimization consistently provides larger improvements, indicating that relay placement is the dominant factor in system performance.}
    \label{fig:ablation}
\end{figure}

Figure~\ref{fig:ablation} shows that both strategies improve performance over the geometric baseline, but relay-aware optimization yields significantly larger gains. On average, relay design improves total distance by approximately \textbf{+19.5\%}, compared to \textbf{+12.6\%} from partition adaptation. Similar trends are observed for delivery efficiency, where relay design provides roughly \textbf{1.5$\times$ greater improvement}.

These results demonstrate that while partition adaptation improves local task allocation, relay design fundamentally reshapes the global transport structure. This validates our central hypothesis that \emph{relay placement is the primary driver of performance in multi-robot delivery systems}.

\subsection{Summary}

Across all evaluated scenarios, VCST-RCP consistently reduces total travel distance and improves delivery efficiency by leveraging relay-based coordination. The gains are most pronounced in large-scale and spatially dispersed environments, where coordinating transport through shared relay structures significantly reduces redundant long-haul motion. In contrast, smaller environments exhibit more modest improvements due to shorter travel distances and higher inherent connectivity. The ablation study further shows that relay-aware optimization provides substantially larger benefits than partition adaptation alone, confirming that relay placement is the dominant factor shaping system performance. Together, these results establish relay-based coordination as a scalable and effective paradigm for multi-robot delivery in open environments.

Looking forward, this work opens several directions for extending relay-based coordination to more realistic settings, including obstacle-rich environments, heterogeneous robot teams, and online task arrivals. Incorporating synchronization-aware scheduling and multi-depot extensions further presents opportunities to bridge the gap between theoretical planning and real-world deployment. More broadly, this work positions relay-based transport as a principled and scalable paradigm for multi-robot coordination in logistics and beyond.

\section{Conclusion}

We presented VCST-RCP, a Voronoi-constrained Steiner tree relay coordination framework that rethinks how multi-robot systems organize transport. Instead of treating deliveries as independent source-to-goal tasks, VCST-RCP explicitly structures package flow through a shared relay backbone, enabling coordinated short-range exchanges that reduce redundant long-haul motion. Across a diverse set of environments, VCST-RCP consistently achieves substantial performance gains, reducing total fleet travel distance by \textbf{31\% on average (up to nearly 50\%)} compared to direct assignment strategies, while delivering over \textbf{50\% higher efficiency} in terms of packages per unit distance. Since locomotion dominates the energy budget of ground robots, these improvements directly translate into meaningful reductions in energy consumption. Importantly, an ablation study reveals that these gains are primarily driven by relay design rather than spatial partitioning, establishing relay placement as the dominant factor governing system performance. These results highlight a fundamental limitation of traditional routing approaches, which optimize individual robot trajectories in isolation but fail to exploit opportunities for coordinated transport. By contrast, VCST-RCP demonstrates that explicitly designing relay structures leads to more scalable and efficient multi-robot systems, particularly in large-scale and spatially distributed environments where coordination is critical.

\bibliographystyle{ieeetr}
\bibliography{mapd}

\end{document}